\documentclass[11pt]{article}

\usepackage[final]{acl}

\usepackage{times}
\usepackage{latexsym}

\usepackage{fontspec}
\usepackage{microtype}

\usepackage{graphicx}
\usepackage{amsmath}
\usepackage{booktabs}
\usepackage{multirow}
\usepackage{caption}
\usepackage{subcaption}
\usepackage{makecell}
\usepackage{array}
\usepackage[frozencache,cachedir=_minted-acl_latex]{minted}
\usepackage{amsfonts,amssymb}

\newcommand{\method}{\textit{DAP}}

\title{Discover and Prove: An Open-source Agentic Framework \\ for Hard Mode Automated Theorem Proving in Lean 4}

\author{
 \textbf{Chengwu Liu\textsuperscript{1}$^*$},
 \textbf{Yichun Yin\textsuperscript{2}},
 \textbf{Ye Yuan\textsuperscript{1}},
 \textbf{Jiaxuan Xie\textsuperscript{3}},
 \textbf{Botao Li\textsuperscript{4}},
 \textbf{Siqi Li\textsuperscript{4}},
\\
 \textbf{Jianhao Shen\textsuperscript{2}},
 \textbf{Yan Xu\textsuperscript{2}},
 \textbf{Lifeng Shang\textsuperscript{2}},
 \textbf{Ming Zhang\textsuperscript{1}},
\\
\\
 \textsuperscript{1}State Key Laboratory for Multimedia Information Processing, School of Computer Science, \\ PKU-Anker LLM Lab, Peking University
 \textsuperscript{2}Huawei Technologies Co., Ltd.\\
 \textsuperscript{3}School of Software \& Microelectronics, Peking University\\
 \textsuperscript{4}School of Electronics Engineering and Computer Science, Peking University
\\
 \small{
   \textbf{Correspondence:} \href{mailto:mzhang_cs@pku.edu.cn}{Ming Zhang (\texttt{mzhang\_cs@pku.edu.cn})}
 }
}

\begin{document}
\maketitle

\def\thefootnote{*}\footnotetext{Work done during the internship at Huawei Technologies Co., Ltd. Code and datasets are available at \url{https://github.com/liuchengwucn/discover-and-prove}.}
\def\thefootnote{\arabic{footnote}}

\begin{abstract}

Most ATP benchmarks embed the final answer within the formal statement — a convention we call ``Easy Mode'' — a design that simplifies the task relative to what human competitors face and may lead to optimistic estimates of model capability.
We call the stricter, more realistic setting ``Hard Mode'': the system must independently discover the answer before constructing a formal proof.
To enable Hard Mode research, we make two contributions.
First, we release MiniF2F-Hard and FIMO-Hard, expert-reannotated Hard Mode variants of two widely-used ATP benchmarks.
Second, we introduce Discover And Prove (\method{}), an agentic framework that uses LLM natural-language reasoning with explicit self-reflection to discover answers, then rewrites Hard Mode statements into Easy Mode ones for existing ATP provers.
\method{} sets the state of the art: on CombiBench it raises solved problems from 7 (previous SOTA, Pass@16) to 10; on PutnamBench it is the first system to formally prove 36 theorems in Hard Mode — while simultaneously revealing that state-of-the-art LLMs exceed 80\% answer accuracy on the same problems where formal provers manage under 10\%, exposing a substantial gap that Hard Mode benchmarks are uniquely suited to measure.
\end{abstract}

\section{Introduction}

The use of AI to solve mathematical problems has attracted considerable research interest, not only because of the potential for concrete applications in domains like education and mathematical research, but also because tackling highly abstract mathematical problems generally requires capabilities that may generalize and transfer to complex real-world tasks.
These capabilities include planning, search, deductive reasoning, and induction. \cite{yang2024formal}
Within the spectrum of mathematical tasks, competition problems --- especially those at the International Mathematical Olympiad (IMO) level --- have garnered particular attention.
These problems go beyond numerical computation or simple formula application.
They typically demand abstraction and modeling, rigorous logical argumentation, and often require elements of intuition and creativity.
Accordingly, the ability to solve IMO-level problems is widely regarded as an important milestone for AI \cite{yang2024formal}.

Existing approaches to solving mathematical problems fall into two broad categories: informal methods and formal methods.
Informal methods solve mathematical problems in natural language and leverage the strong reasoning abilities of large language models (LLMs), whereas formal methods use formal languages such as Lean \cite{moura2021lean} and Isabelle \cite{nipkow2002isabelle} to express the solution.
A key advantage of formal methods is that proofs written in formal languages can be automatically and rigorously verified by a proof assistant program.
At the International Mathematical Olympiad (IMO) 2024, participating AI systems employed formal methods \cite{alphaproof2024ai}.
By IMO 2025, however, most evaluated systems had shifted toward informal approaches \cite{luong2025advanced, huang2025gemini, wei2025aw31, huawei-xiaoyi2025huaweixiaoyi}.

\begin{figure*}[t]
  \centering
  \includegraphics[width=2\columnwidth]{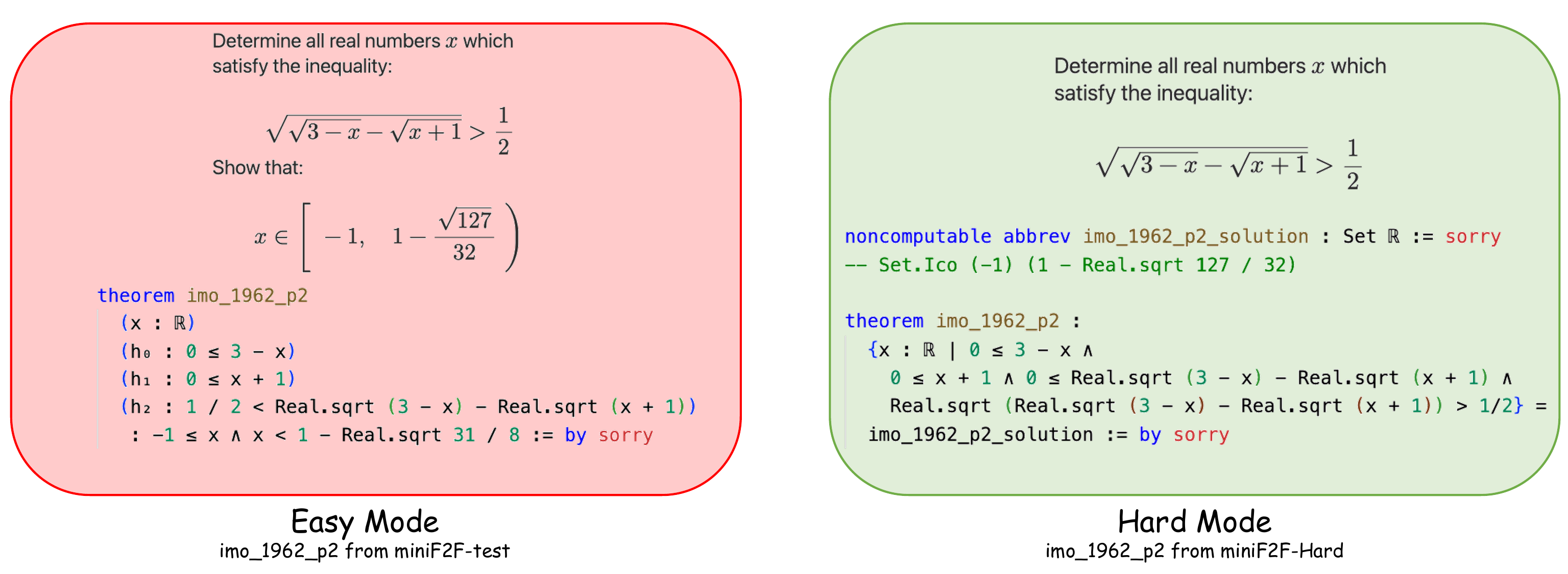}
  \caption{Differences between Easy Mode and Hard Mode configurations in automated theorem proving.
  The example shown is a Lean 4 formalization of an IMO problem in two different styles.
  This example was intentionally selected to illustrate the kind of semantic misalignment that our re-annotation effort corrects: the Easy Mode formalization proves only that $x$ must lie within certain ranges, but does not establish that every value in those ranges is attainable, weakening the original if-and-only-if requirement to a one-directional implication.
  By contrast, our Hard Mode formalization represents the answer as a set, thereby fully capturing the natural-language problem's requirement in the formal statement.}
  \label{fig:easy_hard}
\end{figure*}

We observe that current practice in many formalization efforts often embeds the final answer directly into the statement to be proved, which we refer to as ``Easy Mode''.
We point out that Easy Mode may substantially reduce the difficulty of formal problem-solving tasks.
To address this issue, we draw inspiration from prior works (PutnamBench \cite{tsoukalas2024putnambench} and CombiBench \cite{liu2025combibench}): answer-oriented problems are encoded in Lean 4 with two separate goals (two distinct \texttt{sorry}s). In this ``Hard Mode'' configuration, the model must first supply the final answer by replacing the first \texttt{sorry} with the answer and then produce a conventional formal proof for the remaining goal. This setup prevents embedding extra information that human contestants must discover by themselves in the formal statement.
By definition, \textit{Hard Mode} requires that any quantity a human competitor must derive through reasoning is not supplied as a premise in the formal statement; it must be independently discovered by the AI system.
We adopt the term ``Hard Mode'' following the convention in the Lean community;\footnote{\url{https://leanprover.zulipchat.com/\#narrow/channel/208328-IMO-grand-challenge/topic/IMO.202025.20problem.20statements}}
CombiBench~\cite{liu2025combibench} refers to the same distinction as ``without solution'' vs.\ ``with solution'', while PutnamBench~\cite{tsoukalas2024putnambench} uses ``no answer'' vs.\ ``with answer''.
An example illustrating the difference between Easy Mode and Hard Mode is presented in Figure~\ref{fig:easy_hard}.
We commissioned expert annotators to reannotate two widely used ATP competition datasets, namely MiniF2F and FIMO, producing MiniF2F-Hard and FIMO-Hard.
During reannotation, we corrected known alignment issues in existing formal benchmarks~\cite{wang2025kiminaprover, lin2025goedelproverv2}.

To solve Hard Mode problems, we introduce the Discover and Prove (\method{}) framework, a fully open-source, agent-based ATP framework for Hard Mode ATP tasks.
\method{} consists of two components: a Discovery Module and a Proving Module.
We prompt an open-source LLM to generate and iteratively refine its reasoning and answers using self-verification procedures.
After the Discovery module ``discovers'' a plausible answer and fills the first \texttt{sorry}, the Proving module attempts to produce complete formal proofs by invoking traditional ATP provers.
\method{} achieves state-of-the-art results.
Evaluating on the full PutnamBench dataset, \method{} solves 36 problems in total. On solution-style problems with Hard Mode variants, it solves 19 problems --- to our knowledge, this constitutes the first public result on PutnamBench under this Hard Mode evaluation setting.
On CombiBench Hard Mode, it solves 10 problems, improving significantly on the previous state of the art (Kimina-Prover Preview), which solved 8 problems.

Our contributions are threefold.
\begin{enumerate}
    \item We reannotate two commonly used ATP competition datasets, MiniF2F and FIMO, to align tasks presented to human competitors with those given to AI systems, removing the Easy Mode discrepancy and providing a more principled basis for evaluating AI mathematical capability.
    \item We propose \method{}, an open-source, agentic Hard Mode ATP framework. With a simple and straightforward design, \method{} achieves state-of-the-art performance on PutnamBench and CombiBench.
    \item We provide an analysis quantifying the individual contributions of the two modules of our proposed \method{} prover to overall performance. These results shed light on the relative strengths of informal and formal method approaches for solving competition-level mathematical problems.
\end{enumerate}

\section{Related Work}

\subsection{Mathematical Problem-Solving with AI systems}
Powered by CoT prompting and RLVR training, LLMs have made substantial progress in mathematical reasoning. Frontier models (e.g., OpenAI o1 \cite{jaech2024openai}, DeepSeek R1 \cite{guo2025deepseek}, Google Gemini 2.5 Pro \cite{comanici2025gemini}) now achieve near-saturation performance on widely-used math benchmarks, including GSM8K~\cite{cobbe2021training}, MATH-500~\cite{lightman2023let}, and AIME 2024/2025.

One downside of the informal methods is that the solution traces they produce are notoriously difficult to verify automatically. Assessing the validity of a generated proof would require domain experts to inspect it carefully to detect subtle errors~\cite{lightman2023let}, which is infeasible at scale. For this reason, informal systems are typically applied to problems that require only a final numerical or symbolic answer, verified by direct comparison against ground-truth~\cite{wen2025reinforcement}.

\subsection{Automated Theorem Proving}

The key strength of automated theorem proving (ATP) is that formal proofs can be checked rigorously and automatically by proof assistants~\cite{yousefzadeh2025advocate,wang-etal-2023-dt}. Although formal approaches have faced challenges of limited formal-data availability~\cite{xin2024deepseekproverv15}, they have advanced rapidly with larger LLMs and scaling searching compute~\cite{xin2025bfsprover, chen2025seed}. Recent systems such as Kimina-Prover Preview \cite{wang2025kiminaprover}, DeepSeek-Prover-V2~\cite{ren2025deepseekproverv2}, and Goedel-Prover-V2 \cite{lin2025goedelproverv2} have made substantial progress on MiniF2F \cite{zheng2021minif2f} benchmark.
Seed-Prover \cite{chen2025seed} is a lemma-style whole-proof reasoning model that iteratively refines proofs via Lean compiler feedback, proved lemmas, and self-summarization, achieving over 50\% on PutnamBench and saturating MiniF2F.
DSP~\cite{jiang2022draft} guides formal theorem provers with natural-language draft proofs and structured sketches; DSP+~\cite{cao2025reviving} revives this paradigm with modern reasoning models.
DAP differs from DSP/DSP+ in three key respects; see Appendix~\ref{app:dap-vs-dsp} for a detailed comparison.

Several agentic frameworks have also targeted Lean-based ATP~\cite{thakur2023copra, kumarappan2024leanagent, baba2025proveragent, wang2025malot}; see Appendix~\ref{app:agentic-frameworks}.

To handle informal mathematical problems that require a final answer, prior efforts typically convert problems requiring solutions into proof problems by embedding the desired answer into the formalized statement and proving that statement \cite{zheng2021minif2f, liu2023fimo, xiong-etal-2023-trigo}.
This practice raises two concerns. First, embedding the answer in the statement can reduce the intrinsic difficulty: for many solution-oriented problems, the primary challenge is discovering the answer rather than proving a consequence once known, so supplying it acts as a substantial hint.
Second, the resulting formalized statements are sometimes not perfectly semantically aligned with the tasks human contestants face.
As prior analyses (e.g., FMC~\cite{xie2025fmc}, OlympiadBench~\cite{he-etal-2024-olympiadbench}) have noted, some existing formal benchmarks are misaligned: some formal statements may capture only a subset of the goals that human solvers must address.

\subsection{Formalization \& Data Curation}
\label{subsec:formalization-data}

High-quality formal datasets are scarce and require substantial expert effort~\cite{xin2024deepseekproverv15}.
MiniF2F~\cite{zheng2021minif2f} is one of the most widely used ATP benchmarks, containing formalized statements from mathematical olympiads and high-school and undergraduate courses; Kimina-Prover Preview~\cite{wang2025kiminaprover} supplies corrections to a subset of its problems.
FIMO~\cite{liu2023fimo} is constructed from IMO shortlist problems but lacked a publicly available Lean~4 version, which impeded its adoption in recent work.
PutnamBench~\cite{tsoukalas2024putnambench} comprises hand-constructed formalizations of Putnam Competition problems and, for the first time, provides both ``with answer'' and ``no answer'' evaluation settings.
CombiBench~\cite{liu2025combibench} offers 100 combinatorics problems ranging from middle-school level through IMO and university level, complementing the number-theoretic and algebraic focus of the other benchmarks; IMOSLLean4~\cite{sanjaya2025mortarsanjaya} and IMO-Steps~\cite{yousefzadeh2024lean} additionally supply complete proofs.
To alleviate data scarcity, auto-formalization methods~\cite{wu2022autoformalization} automatically translate informal problems into formal statements.
Representative resources include Lean Workbook~\cite{ying2024lean}, NuminaMath-LEAN~\cite{wang2025kiminaprover}, Goedel-Pset-v1~\cite{lin2025goedelprover}, and FormalMATH~\cite{yuformalmath}.
Because outputs are typically validated only via LLM-as-a-judge~\cite{ying2024lean}, they lack guaranteed semantic correctness and are more commonly used in provers' RL training phases~\cite{wang2025kiminaprover, xin2024deepseekproverv15} than as evaluation benchmarks.
To our knowledge, all auto-formalization techniques follow the Easy Mode paradigm, embedding the desired answer directly into each generated statement.

\section{Methodology}

\begin{figure*}[t]
  \centering
  \includegraphics[width=2\columnwidth]{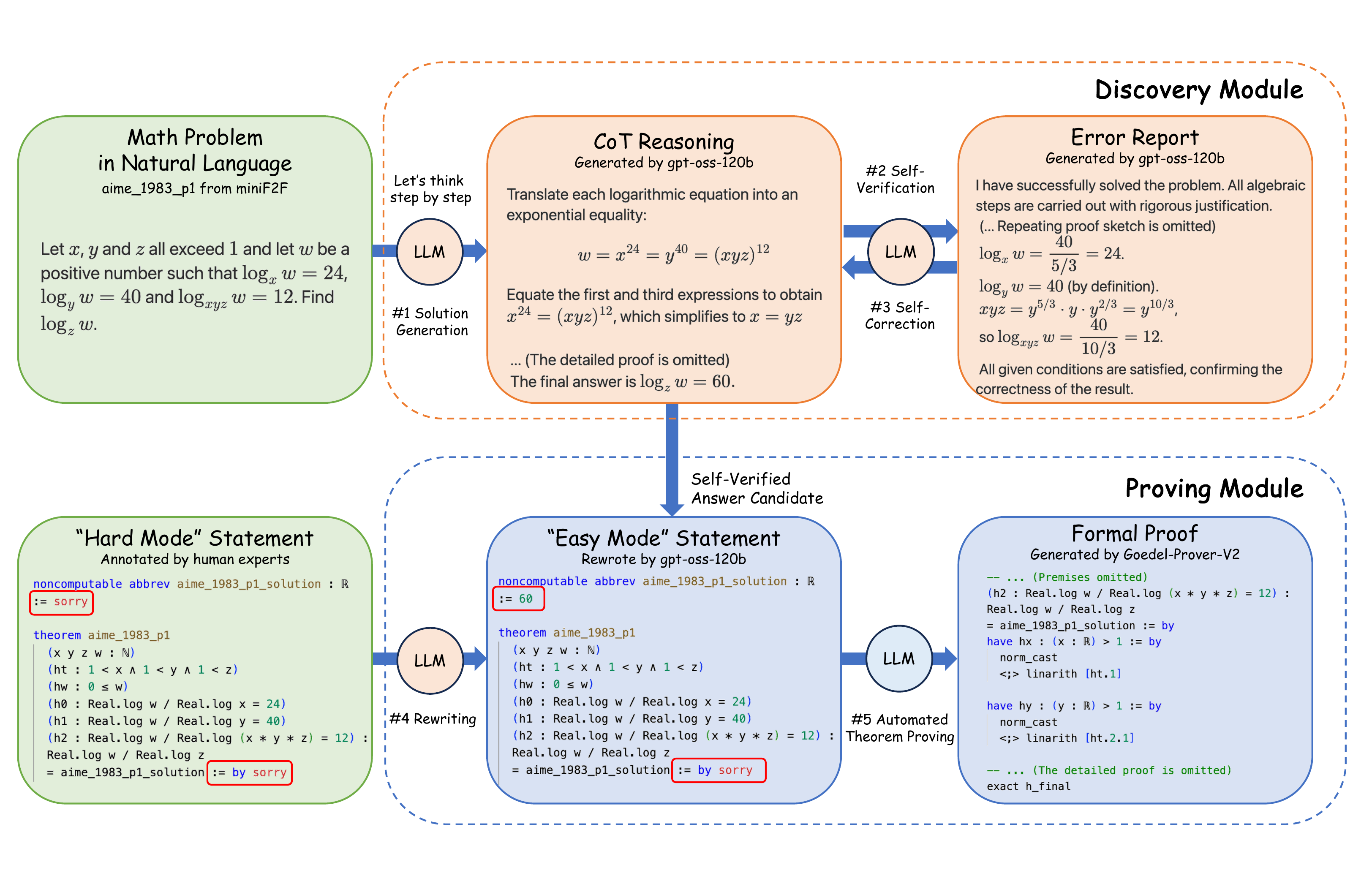}
  \caption{Primary flowchart. A mathematical problem is first processed by the Discovery Module to generate a solution; this solution is then incorporated into the Easy Mode statement during the rewriting stage. Orange circles denote the reasoning LLM, and blue circles denote the theorem prover (another distinct LLM).}
  \label{fig:main_pipeline}
\end{figure*}

To address the challenge of proving Hard Mode Lean 4 theorems containing two \texttt{sorry} placeholders, we propose the Discover and Prove (\method) framework, designed to emulate a human mathematician's approach by providing both the answer and a detailed proof.
As its name suggests, \method{} consists of two primary modules: a Discovery Module and a Proving Module.
The Discovery Module operates in natural language, tasked with identifying the correct solution to the problem, and subsequently transforms the original Hard Mode Lean 4 statement into an Easy Mode statement.
This transformation reduces the number of \texttt{sorry} placeholders from two to one, thereby leaving a single statement that can be resolved by conventional automated theorem provers.
Then, the Proving Module utilizes the Lean 4 formal language to construct a rigorous proof for the solution identified by the Discovery Module.
The complete workflow of the DAP framework is depicted in Figure~\ref{fig:main_pipeline}.

\subsection{Discovery Module}

The Discovery Module aims to solve the original mathematical problem and substitute the solution into the first \texttt{sorry} placeholder of the Lean 4 Hard Mode statement for the Proving Module to use.
This process mirrors human problem-solving by hypothesizing an answer to guide the proof search.
Although the reasoning capabilities of LLMs have substantially improved following the introduction of long Chain-of-Thoughts such as OpenAI o1~\cite{jaech2024openai} and DeepSeek R1~\cite{guo2025deepseek}, the one-shot resolution of highly challenging, IMO-level problems remains unsolved.
Current research indicates that state-of-the-art LLMs often require auxiliary tools (e.g., information retrieval, calculators, external memory) to support deep reasoning~\cite{nakano2022webgptbrowserassistedquestionansweringhuman,luo2025largelanguagemodelagent,yuan2024hybridragcomprehensiveenhancement}.
Inspired by prior work on LLM reasoning systems~\cite{huang2025gemini}, we employ a relatively straightforward configuration where an advanced reasoning model generates solution steps and performs self-verification to enhance accuracy. Specifically, the procedure involves the following steps:

\noindent(1) \textbf{Solution Generation:} Given a mathematical problem in natural language, the model's reasoning capabilities are leveraged to generate a detailed chain-of-thought describing the solution process.\\
(2) \textbf{Self-Verification:} The reasoning LLM is instructed to inspect its steps for potential errors and produce an error report identifying any erroneous locations (a representative error report is shown in Appendix~\ref{app:sv-example}). If self-verification reveals no errors, the process proceeds to step 4; otherwise, it moves to step 3.\\
(3) \textbf{Self-Correction:} The reasoning LLM is instructed to generate a revised solution that addresses the errors identified in the error report.\\
(4) \textbf{Rewriting:} Using the Lean 4 Hard Mode statement, the natural-language problem, and the model's chain-of-thought reasoning, the LLM is prompted to produce a rewritten Lean 4 statement containing only a single placeholder, suitable for automated theorem proving.

All four steps are implemented through meticulously designed prompts to the LLM, which are detailed in Appendix~\ref{app:prompt}. The Discovery Module is crucial because incorrect solutions frequently lead to flawed formal statements that are unprovable from the outset, thus emphasizing the model's capacity for deep reasoning and reliable self-verification.
This framework is released as open-source to facilitate reproducibility and serve as a baseline.
For the Discovery Module, we utilize the open-source model GPT-OSS-120B, known for its strong mathematical reasoning performance.
In principle, the framework can be instantiated with any model that combines robust mathematical reasoning with basic Lean proficiency.

\subsection{Proving Module}

Once the Discovery Module transforms a Hard Mode statement into an Easy Mode formulation, the task becomes a standard ATP problem, amenable to conventional theorem provers.
For this purpose, we employ Goedel-Prover-V2 (32B), a state-of-the-art open-source theorem prover, to process the transformed problems.
By decoupling the reasoning model from the ATP model, the proposed framework can improve as any of the underlying models advance, and provides a contemporary baseline for evaluating LLMs that generate formal mathematical solutions in a proof assistant language.

\section{Data Curation}

\begin{table*}[ht]
    \centering
    \caption{The table presents, for commonly used automated theorem proving datasets, the number of statement samples, the number of Hard Mode problems, the data sources, data curation methods, and compatibility with different versions of the Lean formal language. Notably, ProofNet and miniF2F were originally released as Lean 3 datasets, and publicly available community ports to Lean 4 exist. Although studies have reported performance on a Lean 4 variant of FIMO, no publicly available Lean 4 version of the FIMO dataset exists.}
    \label{tab:data}
    \renewcommand{\arraystretch}{1.2}
\begin{small}
\setlength{\tabcolsep}{2pt}
\begin{tabular}{*{6}{c}}
  \toprule
  & \textbf{\# Samples} & \textbf{\# Hard Mode} & \textbf{Data Source} & \textbf{Curation Method} & \textbf{Compatibility} \\
  \midrule
  ProofNet & 371 & N/A & Textbook & Expert Annotation & Lean 3 \& 4 \\
  miniF2F-test & 244 & N/A & \makecell[c]{Textbook \&\\ Competition} & Expert Annotation & Lean 3 \& 4 \\
  FIMO & 149 & N/A & Competition & Expert Annotation & Lean 3 \\
  PutnamBench & 660 & 340 & Competition & Expert Annotation & Lean 4 \\
  CombiBench & 100 & 45 & \makecell[c]{Textbook \&\\ Competition} & Expert Annotation & Lean 4 \\
  Lean Workbook & 57k & N/A & AoPS Website & Auto-formalization & Lean 4 \\
  NuminaMath-LEAN & 104k & N/A & Competition & \makecell[c]{Auto-formalization \& \\ Expert Annotation} & Lean 4 \\
  \midrule
  \textbf{MiniF2F-Hard} (Ours) & 244 & 194 & \makecell[c]{Textbook \&\\ Competition} & Expert Annotation & Lean 4 \\
  \textbf{FIMO-Hard} (Ours) & 149 & 70 & Competition & Expert Annotation & Lean 4 \\
  \bottomrule
\end{tabular}
\end{small}
    \vspace{-0.5cm}
\end{table*}

\subsection{Annotation Principles}
\label{sec:annotation}

Our re-annotation is guided by three principles.

\noindent\textbf{Semantic Accuracy:} Expert-annotated datasets are small but reliable; auto-formalized datasets are large but unverifiable — current validation methods (LLM-as-a-judge~\cite{ying2024lean}, BEq~\cite{liu2024rethinking}) cannot guarantee semantic alignment with the original natural-language problem.
We therefore start from expert-annotated sources and re-examine each statement manually.

\noindent\textbf{Interpretability:} A formal statement should reflect exactly what a human contestant is asked to do: any quantity to be discovered must not be supplied as a premise, and any claim to be proved must appear as the goal.
Current benchmarks violate this in three recurring ways: (1) encoding the final answer in the statement, (2) weakening the proof goal to a strict subset of the original, or (3) adding premises that human contestants do not have.
Representative instances are shown in Figure~\ref{fig:easy_hard}; we followed the IMOLean~\cite{jsm28} convention throughout.

\noindent\textbf{Consistency:} Formalization style can materially affect ATP success rates~\cite{lin2025goedelprover}, so permitting annotators to choose arbitrary formalizations would introduce evaluation bias.
We provided a unified convention emphasizing idiomatic Lean — ``go further in the direction of idiomatic Lean rather than trying to follow a particular English version closely’’ — consistent with the IMOLean~\cite{jsm28} practice of providing a single canonical statement.

\subsection{Data Selection}

To address these deficiencies, we engaged Lean experts to reannotate two human-expert–annotated datasets, namely MiniF2F~\cite{zheng2021minif2f} and FIMO~\cite{liu2023fimo}. Although ProofNet is also a high-quality, expert-annotated dataset, we observed that all of its natural-language items are inherently proof-based; consequently, we excluded it from consideration. The annotators each have more than one year of Lean-related experience.
The MiniF2F and FIMO datasets were originally produced by experts; we re-examined them in light of error reports documented in the literature~\cite{wang2025kiminaprover} and had each problem independently annotated by two experts to cross-validate labels and thereby safeguard correctness.

\subsection{Dataset Quality Fixes}
\label{sec:quality-fixes}

Beyond creating Hard Mode variants, our annotators performed three non-trivial categories of quality-improvement work; full details are in Appendix~\ref{app:quality-fixes}.
\textbf{Porting FIMO to Lean 4:} we ported all FIMO problems from Lean~3, verifying compilation and semantic faithfulness.
\textbf{Fixing semantic misalignments:} we identified and repaired $\approx$15 errors in miniF2F and $\approx$20 in FIMO across four error types (Table~\ref{tab:misalignment}).
\textbf{Rephrasing for Hard Mode:} unknown values were promoted to free parameters with explicit side-conditions; see Figure~\ref{fig:mathd_algebra_320} (Appendix~\ref{app:annotation-example}).

\section{Experiments}

\begin{table*}[ht]
    \centering
    \caption{Performance of various open-source approaches on standard and Hard Mode benchmarks.
    The numerals beneath each dataset name indicate the total number of examples and the number of Hard Mode examples.
    Best results under each configuration are indicated in boldface.
    In Hard Mode evaluation, each dataset contains proof-style problems (evaluated in their original form) and solution-style problems (evaluated with the answer not provided in the formal statement).
    Each entry $X / Y$ denotes total problems solved ($X$) and solution-style problems solved in Hard Mode ($Y$).
    All results are Pass@32 unless otherwise specified.
    $^\dagger$Kimina-Prover Preview originally reported 7 solved problems on CombiBench at Pass@16; the value of 8 shown here is our re-evaluation at Pass@32.}
    \label{tab:main}
    \renewcommand{\arraystretch}{1.2}
\resizebox{\textwidth}{!}{%
\begin{tabular}{*{6}{c}}
  \toprule
  & & \textbf{PutnamBench} & \textbf{CombiBench} & \textbf{miniF2F-test} & \textbf{FIMO} \\
  & & 660 / 340 & 100 / 45 & 244 / 197 & 149 / 70 \\
  \midrule
  \multirow{5}{*}{\textbf{Easy Mode}} & DeepSeek-Prover-V1.5 & - & 2 & 122 & 1 \\
   & DeepSeek-Prover-V2 (CoT) & 22 & - & 201 & 1 \\
   & Kimina-Prover Preview & - & 8 & 168 & 1 \\
   & Goedel-Prover-SFT & 6 & 3 & 141 & 2 \\
   & Goedel-Prover-V2 & \textbf{43} & \textbf{10} & \textbf{215} & \textbf{4} \\
  \midrule
  \multirow{3}{*}{\textbf{Hard Mode}}
  & Kimina-Prover Preview & - & 8$^\dagger$ / - & - & - \\
  & \textbf{\method{} w/ Agent} (Ours) & \textbf{36 / 19} & 9 / 1 & 201 / 168 & \textbf{3 / 0} \\
  & \textbf{\method{} w/o Agent} (Ours) & 32 / 15 & \textbf{10 / 2} & \textbf{204 / 171} & \textbf{3 / 0} \\
  \bottomrule
\end{tabular}}
    \vspace{-0.5cm}
\end{table*}

For the Discovery Module, we use the open-source model GPT-OSS-120B, which demonstrates strong performance on natural language mathematical reasoning tasks.
For the Proving Module, we use Goedel-Prover-V2, which exhibits state-of-the-art performance in traditional automated theorem proving.
Our formal verification environment uses Lean 4.15.0, with Kimina-Server~\cite{santos2025kimina} mediating interactions with the Lean 4 REPL.

For GPT-OSS-120B, we follow the model’s recommended configuration with sampling temperature 1.0.
During Self-Verification, the model is allowed up to 30 iterative attempts for self-checking and error correction.
If all attempts fail, the pipeline falls back to the no-agent (ablation) configuration, where the reasoning model’s output is used directly as the answer candidate without verification.
For Goedel-Prover-V2, we follow the developer-recommended sampling configuration: temperature 0.7, \texttt{max\_tokens} set to 30,000, 32 samples, and Pass@32 evaluation metric.
We report performance on PutnamBench, CombiBench, miniF2F-Hard, and FIMO-Hard in Table~\ref{tab:main}.

Our method achieves state-of-the-art performance on Hard Mode problems.
On CombiBench, it solves 10 problems, improving on the prior state-of-the-art (8 problems); it is also the first method to solve problems under PutnamBench's ``No Answer'' configuration.
Notably, on PutnamBench and miniF2F-test, our method's Hard Mode accuracy closely approaches Goedel-Prover-V2's Easy Mode performance, suggesting that our Discover and Prove approach effectively reduces Hard Mode problems to Easy Mode problem statements.

\section{Discussion}

\subsection{Ablation Study on Agent Effectiveness}
Advanced reasoning LLMs inherently possess significant self-reflective capabilities~\cite{wen2025reinforcement}. Consequently, explicitly introducing self-verification and self-correction mechanisms might impose an unnecessary reasoning overhead. To investigate when an agentic mode is beneficial, we conducted experiments with the agent mode disabled across the PutnamBench, CombiBench, MiniF2F-Hard, and FIMO-Hard datasets. The results are presented in Table~\ref{tab:main}.

Disabling the agent's explicit self-verification and self-correction mechanisms resulted in a significant performance degradation on challenging datasets like PutnamBench.
Conversely, no performance degradation was observed on lower-difficulty datasets, such as CombiBench and MiniF2F.
We hypothesize that this effect stems from these datasets containing a substantial proportion of relatively simple problems (e.g., middle-school textbook problems) that reasoning LLMs can solve with minimal difficulty. In such cases, the explicit inclusion of self-verification might inadvertently cause the model's instruction-following behavior to over-dominate, leading to excessive self-questioning and the introduction of additional noise to the solution.

\begin{table}[ht]
    \centering
    \caption{On the subset of Hard Mode problems with definitive (ground-truth) answers, the Discovery Module's performance in solving mathematical problems presented in natural language was evaluated.}
    \label{tab:ablation}
\begin{footnotesize}
\resizebox{\linewidth}{!}{
\begin{tabular}{lccc}
  \toprule
  \textbf{Benchmark}    & \textbf{Size} & \textbf{\method{} w/ Agent} & \textbf{\method{} w/o Agent} \\
  \midrule
  PutnamBench  & 340  & 293 (86\%) & 265 (78\%) \\
  CombiBench   & 45   & 32 (71\%) & 29 (64\%) \\
  miniF2F-Hard & 197  & 197 (100\%) & 197 (100\%) \\
  FIMO-Hard    & 70   & 43 (61\%) & 38 (54\%) \\
  \bottomrule
\end{tabular}
}
\end{footnotesize}
    \vspace{-0.5cm}
\end{table}

We analyzed the Discovery Module's accuracy on Hard Mode problems with ground-truth answers, which permit direct assessment of natural language responses (Table~\ref{tab:ablation}).
Even without agentic self-reflection, the model achieves approximately 78\% correctness on Putnam problems, with performance on MiniF2F approaching saturation.
For lower-difficulty problems, explicit self-verification and self-correction can be superfluous given the already high one-shot accuracy of reasoning LLMs.
However, for more challenging benchmarks like PutnamBench and FIMO-Hard, these agentic components remain crucial.
A fine-grained failure-mode analysis of the Discovery Module on CombiBench and FIMO is provided in Appendix~\ref{app:failure}.

Ablating self-verification iteration counts shows that 10 iterations approach saturation and 30 (our default) add only marginal gains; see Appendix~\ref{app:sv-ablation}.

\subsection{Ablation on Rewriting Strategies}
\label{sec:ablation-rewriting}

A central design choice in \method{} is the two-stage rewriting pipeline: the Discovery Module first derives the answer in natural language, and only then is the Hard Mode statement transformed into an Easy Mode statement for the ATP prover.
To justify this design, we compare three alternative strategies.
\textbf{No Rewriting:} The Hard Mode problem statement (containing two \texttt{sorry} placeholders) is fed directly to the ATP prover without any rewriting.
\textbf{Straight Rewriting:} The Discovery Module is discarded; instead, an LLM is asked to simultaneously discover the answer \emph{and} produce the rewritten Easy Mode statement in a single step.
\textbf{Proposed Rewriting (Ours):} The Discovery Module first finds the answer; the Rewriting stage then transforms the statement using that answer.

\begin{table}[ht]
    \centering
    \caption{Pass@32 comparison of three rewriting strategies.
    No Rewriting and Straight Rewriting results were manually verified to exclude spurious proofs.}
    \label{tab:ablation-rewriting}
\resizebox{\linewidth}{!}{
\begin{tabular}{lccc}
  \toprule
  \textbf{Benchmark} & \textbf{No} & \textbf{Straight} & \textbf{Ours} \\
  \midrule
  PutnamBench (660)   & 23 & 19 & \textbf{36} \\
  CombiBench (100)    &  \textbf{9} &  \textbf{9} &  \textbf{9} \\
  miniF2F-Hard (244)  & 57 & 38 & \textbf{201} \\
  FIMO-Hard (149)     &  \textbf{3} &  \textbf{3} &  \textbf{3} \\
  \bottomrule
\end{tabular}
}
    \vspace{-0.3cm}
\end{table}

Table~\ref{tab:ablation-rewriting} shows the results, from which we draw three key observations.

\paragraph{No Rewriting is largely ineffective on Hard Mode problems.}
We attribute this to models' lack of training on Hard Mode ATP tasks, producing large out-of-distribution performance drops when asked to both discover answers and prove them directly in the formal system.

\paragraph{Straight Rewriting performs no better — and sometimes worse.}
Requiring the model to handle answer discovery and formalization simultaneously in a single step causes it to frequently fail at both.
The joint burden of informal mathematical reasoning and syntactically precise Lean code generation leads to degraded performance on most benchmarks.

\paragraph{Spurious proofs explain the remaining gap.}
We frequently observed \emph{spurious proofs} (cheating behavior) in both No Rewriting and Straight Rewriting settings.
Because the model can see the full Lean statement — including the \texttt{abbrev solution} definition — during solving, it sometimes avoids genuine mathematical reasoning by copying problem conditions directly into the proof.
A concrete example from \texttt{fimo\_2009\_algebra\_p3} is shown in Figure~\ref{fig:spurious_proof} (Appendix~\ref{app:spurious-proof}).
\method{}'s decoupled design prevents this shortcut: the prover only ever receives the rewritten Easy Mode statement, which does not expose the answer placeholder, eliminating the opportunity for such cheating behavior.

\method{}'s modular design allows the Discovery Module and the Proving Module to be replaced independently; experiments with both lightweight small open-source models and the stronger Aristotle API confirm that the pipeline remains functional across resource constraints and that stronger informal reasoning models yield further gains (see Appendix~\ref{app:cross-model}).

\subsection{Natural Language Reasoning or Formal Language Reasoning?}
The choice between natural-language and formal-language reasoning presents a critical dilemma for AI's advancement in complex problem-solving.
Recent innovations in natural-language reasoning---such as Long Chain-of-Thought and Reinforcement Learning from Human Feedback (RLHF), exemplified by OpenAI's o1 and DeepSeek's R1---have significantly enhanced LLMs' capabilities.
Research has shown that natural-language representations can exhibit very high empirical ceilings; performance on challenging datasets like AIME has saturated rapidly~\cite{jaech2024openai, wen2025reinforcement}.
At IMO 2025, the majority of AI systems transitioned from formal to natural-language representations and successfully solved creative reasoning problems, with several solutions attaining gold-medal recognition.
In contrast, formal-language systems like Seed Prover~\cite{chen2025seed} and Aristotle~\cite{achim2025aristotle} achieved medal-worthy results through extensive computational search, with some solutions completed only after official submission deadlines.

This study investigates Hard Mode ATP as a neutral benchmark for comparing natural-language and formal-language reasoning.
We observe a similar divergence: state-of-the-art LLMs with self-reflection exceed 80\% accuracy on PutnamBench, while formal-language systems achieve less than 10%

Despite the robust empirical performance of natural-language reasoning, formal mathematical language will remain essential for future AI systems in mathematical reasoning.
Natural-language proofs are prone to subtle errors that are challenging to identify without formal verification, significantly limiting their practical utility for critical applications such as education and advanced mathematics research.
As evidenced by results on datasets like Putnam and FIMO, formal mathematical reasoning still requires considerable development.
We therefore propose that a promising avenue for future research involves integrating the strengths of both natural-language and formal-language approaches to bridge the existing performance disparity.

\section{Conclusion}

This study identifies semantic accuracy, interpretability, and transparency problems in current Easy Mode automated theorem-proving settings.
We argue that formal statements in Hard Mode, manually annotated by human experts with a unified convention, more accurately reflect the problem-solving requirements faced by participants in mathematical competitions.

To address the scarcity of Hard Mode ATP benchmarks, we engaged Lean 4 experts to reannotate high-quality datasets into the Hard Mode format, producing MiniF2F-Hard and FIMO-Hard.
In parallel, we rectified existing semantic misalignments and enforced a unified convention.

To address Hard Mode problems, we introduce the \method{} framework, designed to offer a rigorous evaluation of AI systems' capacity to solve complex formal mathematical problems.
The framework achieves state-of-the-art performance on PutnamBench and CombiBench, representing the first reported progress on PutnamBench's Hard Mode configuration.
These results demonstrate the efficacy of combining natural-language reasoning to derive answers with formal methods to prove them.

\section*{Limitations}

The primary limitation of this work concerns potential dataset contamination, which may have led to inflated estimates of reported AI performance.
The miniF2F and FIMO datasets annotated in this study, together with their source problems in natural language (like IMO shortlisted problems) and other related datasets, are publicly accessible on the Internet.
The two main LLMs employed in this work---GPT-OSS-120B and Goedel-Prover-V2---have released model weights but not their training corpora; consequently, we cannot definitively determine whether these datasets were included in training data.

By demonstrating a performance gap between the Discovery Module and the Proving Module, this study identifies the bottleneck for Hard Mode ATP as formal reasoning rather than natural-language reasoning.
This suggests that approaches leveraging natural-language reasoning to assist formal-language reasoning, such as DSP~\cite{jiang2022draft}, show promise.
In our work, however, the natural-language output passed to the Proving Module is limited to answer candidates.
The integration between formal-language and natural-language reasoning therefore remains limited.
Ideally, the two modules would interact more tightly, with one modality providing assistance when the other encounters difficulty. Developing such tighter, cooperative integration is left to future work.

\section*{Acknowledgments}

This paper is partially supported by grants from the National Key Research and Development Program of China with Grant No. 2023YFC3341203.

\bibliography{custom}

\newpage
\appendix
\section{Agentic Frameworks for Formal Theorem Proving}
\label{app:agentic-frameworks}

Beyond pure proof-search approaches, several agentic frameworks have been explored for Lean-based ATP.
COPRA~\cite{thakur2023copra} converts a general-purpose LLM into a Lean proof specialist via a language-agent loop.
LeanAgent~\cite{kumarappan2024leanagent} studies lifelong learning to continuously improve LLM performance on advanced mathematics.
ProverAgent~\cite{baba2025proveragent} leverages non-formal models to propose auxiliary lemmas that guide formal proofs.
MA-LoT~\cite{wang2025malot} applies a multi-agent, Lean-based long chain-of-thought approach with iterative proof repair via Lean compiler feedback.
None of these systems are directly compared to \method{} experimentally, as they target standard (Easy Mode) ATP tasks and their setups are not directly comparable.

\section{Comparison between DAP and DSP/DSP+}
\label{app:dap-vs-dsp}

DAP differs from DSP/DSP+ in three key respects.
First, \textbf{target task}: DSP/DSP+ address standard ATP where the full formal statement is already given; DAP targets Hard Mode ATP, where the system must first discover the answer before proving.
Second, \textbf{inter-module communication}: DSP passes intermediate reasoning steps (draft and proof sketch) to the formal prover; DAP passes only the final answer to the rewriting stage. Passing only the answer avoids reliance on particular formal-language idioms and sidesteps the problem that natural-language solution steps are often awkward or counterproductive when injected directly into tactic-based Lean proofs~\cite{liu2023fimo}. Additionally, DSP's design is tightly coupled to Isabelle's subgoal syntax, making it difficult to port to Lean, while DSP+'s sketch steps require single equations expressible as Lean \texttt{have} statements, which limits applicability.
Third, \textbf{verification design}: DAP front-loads self-verification inside the Discovery Module before passing to the prover; DSP includes no self-verification, and DSP+ performs repair only when the formal sketch has a syntactic error.

\section{Prompts used in Discovery Module}
\label{app:prompt}

\subsection{Prompt for Solution Generation}
\begin{minted}{text}
### Core Instructions ###

*   **Rigor is Paramount:** Your primary goal is to produce a complete and rigorously justified solution. Every step in your solution must be logically sound and clearly explained. A correct final answer derived from flawed or incomplete reasoning is considered a failure.
*   **Honesty About Completeness:** If you cannot find a complete solution, you must **not** guess or create a solution that appears correct but contains hidden flaws or justification gaps. Instead, you should present only significant partial results that you can rigorously prove. A partial result is considered significant if it represents a substantial advancement toward a full solution. Examples include:
    *   Proving a key lemma.
    *   Fully resolving one or more cases within a logically sound case-based proof.
    *   Establishing a critical property of the mathematical objects in the problem.
    *   For an optimization problem, proving an upper or lower bound without proving that this bound is achievable.
*   **Use TeX for All Mathematics:** All mathematical variables, expressions, and relations must be enclosed in TeX delimiters (e.g., `Let $n$ be an integer.`).

### Output Format ###

Your response MUST be structured into the following sections, in this exact order.

**1. Summary**

Provide a concise overview of your findings. This section must contain two parts:

*   **a. Verdict:** State clearly whether you have found a complete solution or a partial solution.
    *   **For a complete solution:** State the final answer, e.g., "I have successfully solved the problem. The final answer is..."
    *   **For a partial solution:** State the main rigorous conclusion(s) you were able to prove, e.g., "I have not found a complete solution, but I have rigorously proven that..."
*   **b. Method Sketch:** Present a high-level, conceptual outline of your solution. This sketch should allow an expert to understand the logical flow of your argument without reading the full detail. It should include:
    *   A narrative of your overall strategy.
    *   The full and precise mathematical statements of any key lemmas or major intermediate results.
    *   If applicable, describe any key constructions or case splits that form the backbone of your argument.

**2. Detailed Solution**

Present the full, step-by-step mathematical proof. Each step must be logically justified and clearly explained. The level of detail should be sufficient for an expert to verify the correctness of your reasoning without needing to fill in any gaps. This section must contain ONLY the complete, rigorous proof, free of any internal commentary, alternative approaches, or failed attempts.

### Self-Correction Instruction ###

Before finalizing your output, carefully review your "Method Sketch" and "Detailed Solution" to ensure they are clean, rigorous, and strictly adhere to all instructions provided above. Verify that every statement contributes directly to the final, coherent mathematical argument.
\end{minted}

\subsection{Prompt for Self-Verification}
\begin{minted}{text}
Below is the bug report. If you agree with certain item in it, can you improve your solution so that it is complete and rigorous? Note that the evaluator who generates the bug report can misunderstand your solution and thus make mistakes. If you do not agree with certain item in the bug report, please add some detailed explanations to avoid such misunderstanding. Your new solution should strictly follow the instructions in the system prompt.
\end{minted}

\subsection{Prompt for Self-Correction}
\begin{minted}{text}
You are an expert AI assistant specializing in Formal Mathematics with the Lean 4 proof assistant. Your task is to perform a specific and targeted code modification.

You will be given three inputs:
1.  A natural language math problem (`<NATURAL_LANGUAGE_PROBLEM>`).
2.  A detailed natural language solution to that problem (`<NATURAL_LANGUAGE_SOLUTION>`).
3.  A Lean 4 code statement that formalizes the problem (`<LEAN4_STATEMENT>`).

The Lean 4 statement contains a "fill-in-the-blank" definition for the solution, typically an `abbrev` or `def` with `sorry` as its value.

**Your task is to:**

1.  **Read the `<NATURAL_LANGUAGE_SOLUTION>` to find the final answer** to the problem. The answer is usually explicitly stated at the end (e.g., "The value is π/2", "The result is 42").
2.  **Locate the `abbrev` or `def` line** in the `<LEAN4_STATEMENT>` that defines the solution variable (e.g., `abbrev my_solution : ℝ := sorry`).
3.  **Replace the `sorry`** in that definition with the final answer you extracted. Ensure the syntax is correct for Lean 4 (e.g., `π` becomes `Real.pi`).
4.  **Output the entire Lean 4 code block with this single modification.** Do not change any other part of the code. Specifically, the `theorem`'s proof, which is also `sorry`, must remain unchanged.

**Example:**

**<NATURAL_LANGUAGE_PROBLEM>:**
Evaluate the integral \[\int_0^1 \frac{1}{1+x^2} \,dx.\]

**<NATURAL_LANGUAGE_SOLUTION>:**
The problem is to evaluate the definite integral of \(f(x) = \frac{1}{1+x^2}\) from \(0\) to \(1\).

The antiderivative of \(\frac{1}{1+x^2}\) is \(\arctan(x)\).
Using the Fundamental Theorem of Calculus, we have:
\[
\int_0^1 \frac{1}{1+x^2} \,dx = [\arctan(x)]_0^1
\]
\[
= \arctan(1) - \arctan(0)
\]
Since \(\arctan(1) = \pi/4\) and \(\arctan(0) = 0\), the value of the integral is:
\[
\frac{\pi}{4} - 0 = \frac{\pi}{4}.
\]
The final answer is \(\pi/4\).

**<LEAN4_STATEMENT>:**
```lean
import Mathlib.Analysis.SpecialFunctions.Trigonometric.Arctan

open Set

abbrev integral_value : ℝ := sorry

theorem integral_example
  : ∫ x in Icc 0 1, 1 / (1 + x^2) = integral_value :=
sorry
```

**Expected Output:**
```lean
import Mathlib.Analysis.SpecialFunctions.Trigonometric.Arctan

open Set

abbrev integral_value : ℝ := Real.pi / 4

theorem integral_example
  : ∫ x in Icc 0 1, 1 / (1 + x^2) = integral_value :=
sorry
```

---

Now, perform the task for the following inputs.

**<NATURAL_LANGUAGE_PROBLEM>:**
{{NATURAL_LANGUAGE_PROBLEM}}

**<NATURAL_LANGUAGE_SOLUTION>:**
{{NATURAL_LANGUAGE_SOLUTION}}

**<LEAN4_STATEMENT>:**
```lean
{{LEAN4_STATEMENT}}
```
\end{minted}

\subsection{Prompt for Rewriting}
\begin{minted}{text}
You are an expert mathematician and a meticulous grader for an International Mathematical Olympiad (IMO) level exam. Your primary task is to rigorously verify the provided mathematical solution. A solution is to be judged correct **only if every step is rigorously justified.** A solution that arrives at a correct final answer through flawed reasoning, educated guesses, or with gaps in its arguments must be flagged as incorrect or incomplete.

### Instructions ###

**1. Core Instructions**
*   Your sole task is to find and report all issues in the provided solution. You must act as a **verifier**, NOT a solver. **Do NOT attempt to correct the errors or fill the gaps you find.**
*   You must perform a **step-by-step** check of the entire solution. This analysis will be presented in a **Detailed Verification Log**, where you justify your assessment of each step: for correct steps, a brief justification suffices; for steps with errors or gaps, you must provide a detailed explanation.

**2. How to Handle Issues in the Solution**
When you identify an issue in a step, you MUST first classify it into one of the following two categories and then follow the specified procedure.

*   **a. Critical Error:**
    This is any error that breaks the logical chain of the proof. This includes both **logical fallacies** (e.g., claiming that `A>B, C>D` implies `A-C>B-D`) and **factual errors** (e.g., a calculation error like `2+3=6`).
    *   **Procedure:**
        *   Explain the specific error and state that it **invalidates the current line of reasoning**.
        *   Do NOT check any further steps that rely on this error.
        *   You MUST, however, scan the rest of the solution to identify and verify any fully independent parts. For example, if a proof is split into multiple cases, an error in one case does not prevent you from checking the other cases.

*   **b. Justification Gap:**
    This is for steps where the conclusion may be correct, but the provided argument is incomplete, hand-wavy, or lacks sufficient rigor.
    *   **Procedure:**
        *   Explain the gap in the justification.
        *   State that you will **assume the step's conclusion is true** for the sake of argument.
        *   Then, proceed to verify all subsequent steps to check if the remainder of the argument is sound.

**3. Output Format**
Your response MUST be structured into two main sections: a **Summary** followed by the **Detailed Verification Log**.

*   **a. Summary**
    This section MUST be at the very beginning of your response. It must contain two components:
    *   **Final Verdict**: A single, clear sentence declaring the overall validity of the solution. For example: "The solution is correct," "The solution contains a Critical Error and is therefore invalid," or "The solution's approach is viable but contains several Justification Gaps."
    *   **List of Findings**: A bulleted list that summarizes **every** issue you discovered. For each finding, you must provide:
        *   **Location:** A direct quote of the key phrase or equation where the issue occurs.
        *   **Issue:** A brief description of the problem and its classification (**Critical Error** or **Justification Gap**).

*   **b. Detailed Verification Log**
    Following the summary, provide the full, step-by-step verification log as defined in the Core Instructions. When you refer to a specific part of the solution, **quote the relevant text** to make your reference clear before providing your detailed analysis of that part.

**Example of the Required Summary Format**
*This is a generic example to illustrate the required format. Your findings must be based on the actual solution provided below.*

**Final Verdict:** The solution is **invalid** because it contains a Critical Error.

**List of Findings:**
*   **Location:** "By interchanging the limit and the integral, we get..."
    *   **Issue:** Justification Gap - The solution interchanges a limit and an integral without providing justification, such as proving uniform convergence.
*   **Location:** "From $A > B$ and $C > D$, it follows that $A-C > B-D$"
    *   **Issue:** Critical Error - This step is a logical fallacy. Subtracting inequalities in this manner is not a valid mathematical operation.
\end{minted}

\subsection{Example of Self-Verification Error Report}
\label{app:sv-example}

The Self-Verification step prompts the LLM to carefully review the generated solution step by step and produce a structured error report.
Each finding identifies a specific location in the solution, describes the issue, and classifies its severity.
The following is a representative example of such an error report, produced by the Discovery Module on a functional-equation problem.

\begin{minted}{text}
Final Verdict: The solution contains several Justification Gaps and therefore is not fully rigorous.

List of Findings:
* Location: Lemma 1 proof -- "h(x0) is an upper bound of the set {y | y < x0} ..."
  Issue: Justification Gap -- the claim that h(x0) bounds all predecessors of x0 is not proved; the argument that this contradicts the supremum property is invalid.
* Location: Lemma 2 proof -- "Both lines are proved by a simple induction on the natural variable."
  Issue: Justification Gap -- the induction for the second coordinate is only sketched; it is not shown that the composed beta-functions satisfy the required monotonicity and domination properties.
* Location: Lemma 2 proof -- base case "For n=0 ..." while N was earlier taken to start at 1.
  Issue: Minor Justification Gap -- the indexing mismatch is not addressed, though it does not affect the overall argument.
\end{minted}

\section{Dataset Quality Fixes}
\label{app:quality-fixes}

\noindent\textbf{Porting FIMO to Lean 4:}
The publicly available FIMO dataset is written in Lean~3.
Although a largely automated migration tool exists, the syntactic and library-level changes between Lean~3 and Lean~4 mean that many statements do not port cleanly: tactic names changed, Mathlib APIs were reorganised, and some constructs require manual rewriting to compile while remaining semantically faithful.
Our annotators ported every FIMO problem to Lean~4, verifying that each statement compiles and is semantically faithful to the original.

\noindent\textbf{Fixing Semantic Misalignments:}
Following the annotation principles described in \S\ref{sec:annotation}, annotators identified and repaired formalization errors present in the source datasets.
We found approximately 15 misalignments in miniF2F and 20 in FIMO.
Table~\ref{tab:misalignment} summarises the four most common error types.

\begin{table*}[ht]
\centering
\caption{Semantic misalignment types found and corrected in miniF2F and FIMO during our re-annotation.}
\label{tab:misalignment}
\renewcommand{\arraystretch}{1.2}
\begin{small}
\setlength{\tabcolsep}{4pt}
\begin{tabular}{m{4.5cm}ccm{6.0cm}}
\toprule
\textbf{Misalignment Type} & \textbf{\# in miniF2F} & \textbf{\# in FIMO} & \textbf{Example} \\
\midrule
Ignoring whether an extremum is attainable & 8 & 8 &
  \texttt{mathd\_numbertheory\_495}, \texttt{fimo\_2010\_number\_theory\_p1\_1} \\
Adding extra conditions in the formalization & 2 & 1 &
  \texttt{mathd\_algebra\_320}, \texttt{fimo\_2017\_number\_theory\_p8} \\
Incomplete proof goals & 2 & 2 &
  \texttt{amc12b\_2021\_p3}, \texttt{fimo\_2008\_algebra\_p1} \\
Missing/incomplete NL problem statements & 2 & 2 &
  \texttt{mathd\_algebra\_188}, \texttt{fimo\_2008\_algebra\_p3\_1} \\
\bottomrule
\end{tabular}
\end{small}
\end{table*}

\noindent\textbf{Rephrasing for Hard Mode Compatibility:}
Our annotation principle requires that any quantity a human competitor must derive must not be hard-coded in the formal statement.
In some cases this meant non-trivially rephrasing a problem so that the unknown value becomes a free parameter and proper side-conditions (simplicity, squarefreeness, positivity, etc.) are added explicitly.
A representative before/after example is shown in Figure~\ref{fig:mathd_algebra_320} (Appendix~\ref{app:annotation-example}).

\section{Hard Mode Annotation Example}
\label{app:annotation-example}

Figure~\ref{fig:mathd_algebra_320} shows a representative before/after example of rephrasing a problem for Hard Mode compatibility.
The original formalization hard-codes \texttt{c\,=\,2}, leaking part of the answer; the rephrased version promotes \texttt{c} to a free parameter and adds explicit simplicity and squarefreeness conditions, requiring the solver to determine the canonical form independently.

\begin{figure*}[t]
\centering
\begin{minipage}{\textwidth}
\noindent\textit{Natural-language problem: Let $x$ be a positive number such that $2x^2 = 4x + 9$. If $x$ can be written in simplified form as $\dfrac{a + \sqrt{b}}{c}$ where $a$, $b$, and $c$ are positive integers, what is $a + b + c$?}

\medskip
\noindent\textit{Original formalization (problematic) — \texttt{c\,=\,2} is hard-coded:}
\begin{minted}[fontsize=\footnotesize, breaklines, frame=single]{lean4}
theorem mathd_algebra_320
    (x : ℝ) (a b c : ℕ)
    (h0 : 0 < a ∧ 0 < b ∧ 0 < c ∧ 0 ≤ x)
    (h1 : 2 * x ^ 2 = 4 * x + 9)
    (h2 : x = (a + Real.sqrt b) / c)
    (h3 : c = 2) : a + b + c = 26 := by
\end{minted}

\noindent\textit{Rephrased formalization (aligned) — \texttt{c} is a free parameter:}
\begin{minted}[fontsize=\footnotesize, breaklines, frame=single]{lean4}
abbrev mathd_algebra_320_solution : ℕ := sorry
-- 26

theorem mathd_algebra_320
    (x : ℝ) (a b c : ℕ)
    (h_x_pos : 0 < x)
    (h_eqn : 2 * x ^ 2 = 4 * x + 9)
    (h_form : x = (a + Real.sqrt b) / c)
    (h_abc_pos : a > 0 ∧ b > 0 ∧ c > 0)
    (h_simplified_gcd : Nat.gcd a c = 1)
    (h_simplified_sq_free : Squarefree b)
    : a + b + c = mathd_algebra_320_solution := by
      sorry
\end{minted}
\end{minipage}
\caption{Rephrasing \texttt{mathd\_algebra\_320} for Hard Mode compatibility. The original formalization hard-codes \texttt{c\,=\,2}, leaking part of the answer; the rephrased version promotes \texttt{c} to a free parameter and adds explicit simplicity and squarefreeness conditions, requiring the solver to determine the canonical form independently.}
\label{fig:mathd_algebra_320}
\end{figure*}

\section{Failure-Mode Analysis of the Discovery Module}
\label{app:failure}

To understand the limitations of the Discovery Module, we analyzed all Discovery outputs on the 45 solution-style problems from CombiBench and the 70 solution-style problems from FIMO.
These two sets are moderate in difficulty (avoiding the saturation seen on miniF2F-Hard) and cover a broad range of competition mathematics areas (combinatorics, number theory, and algebra), making them well-suited for diagnosing real weaknesses.
We manually categorized every Discovery failure and summarize the main error types in Table~\ref{tab:failure}.

\begin{table*}[t]
\centering
\footnotesize
\caption{Failure-mode analysis of the Discovery Module on FIMO and CombiBench, with and without self-verification (SV). Counts indicate the number of problems exhibiting each failure type.}
\label{tab:failure}
{\setlength{\tabcolsep}{3pt}%
\begin{tabular}{m{4.0cm}ccccm{5.8cm}}
\toprule
\multirow{2}{*}{\textbf{Error Type}} & \textbf{FIMO} & \textbf{FIMO} & \textbf{Combi} & \textbf{Combi} & \multirow{2}{*}{\textbf{Example}} \\
 & \textbf{w/o SV} & \textbf{w/ SV} & \textbf{w/o SV} & \textbf{w/ SV} & \\
\midrule
Failed to correctly understand the problem        & 0 & 0 & 3 & 3 & \texttt{imo\_2010\_p5}: invented a non-existent operation \\
Ignored or misused constraints                    & 3 & 2 & 2 & 2 & \texttt{fimo\_2010\_number\_\allowbreak theory\_p1\_2}: omitted case analysis \\
Incorrect mathematical reasoning foundations      & 2 & 2 & 5 & 3 & \texttt{fimo\_2016\_algebra\_p3}: induction without base case \\
Minor arithmetic / computational error            & 9 & 4 & 0 & 0 & \texttt{fimo\_2012\_algebra\_p1}: arithmetic mistake \\
Insufficient depth / beyond model capability      & 4 & 2 & 2 & 2 & \texttt{fimo\_2013\_algebra\_p6}: missing preprocessing \\
\bottomrule
\end{tabular}}
\end{table*}

\paragraph{FIMO: high difficulty in number theory and algebra.}
Many FIMO problems remain beyond the model's capabilities; self-verification cannot fully recover correct answers for the hardest instances.
Function-equation problems are particularly challenging and often fail even with verification, as the underlying reasoning requires multi-step algebraic manipulation that is difficult to check automatically.

\paragraph{CombiBench: comprehension failures.}
On CombiBench, we frequently observed that the model did not correctly understand problem statements or even hallucinated conditions that do not appear in the original problem.
These comprehension failures are not substantially reduced by self-verification.
We hypothesize that certain combinatorial phrasings---though natural to humans---are less accessible to current LLMs and lead to misparsing or misinterpretation.

\paragraph{Self-verification helps but is not a cure-all.}
For number-theory and algebra problems, self-verification notably reduces minor computational errors (9~$\to$~4) and some shallow-reasoning failures (4~$\to$~2), rescuing answers lost due to arithmetic mistakes.
For combinatorics, self-verification helps reduce instances of incorrect reasoning foundations (5~$\to$~3) but does not eliminate comprehension or hallucination errors.
Overall, self-verification is most effective when the error is a localized mistake (e.g., an arithmetic slip) rather than a fundamental misunderstanding of the problem.

\section{Self-Verification Iteration Ablation}
\label{app:sv-ablation}

Table~\ref{tab:ablation-sv} quantifies the effect of the maximum number of self-verification iterations on Discovery Module accuracy.

\begin{table}[ht]
    \centering
    \caption{Discovery Module accuracy (number of correctly solved problems) under varying maximum self-verification iteration budgets. Column headers indicate the iteration limit; 0 means no self-verification. $^\dagger$Our default setting. All results are Pass@32.}
    \label{tab:ablation-sv}
    \small
    {\setlength{\tabcolsep}{4pt}%
    \begin{tabular}{lcccc}
      \toprule
      & \multicolumn{4}{c}{\textbf{Max SV Iterations}} \\
      \cmidrule(lr){2-5}
      \textbf{Dataset} & \textbf{0} & \textbf{5} & \textbf{10} & \textbf{30}$^\dagger$ \\
      \midrule
      PutnamBench (340)  & 265 & 291 & 292 & \textbf{293} \\
      CombiBench (45)    &  29 &  31 &  \textbf{32} & \textbf{32} \\
      miniF2F-Hard (194) & \textbf{194} & \textbf{194} & \textbf{194} & \textbf{194} \\
      FIMO-Hard (70)     &  38 &  40 &  \textbf{43} & \textbf{43} \\
      \bottomrule
    \end{tabular}}
\end{table}

Two observations emerge.
First, on miniF2F-Hard the Discovery Module achieves perfect accuracy even without self-verification, consistent with the saturation effect (though possible dataset contamination should be noted).
Second, for harder benchmarks like PutnamBench and FIMO-Hard, self-verification provides substantial gains; 10 iterations approach saturation and balance computational cost with accuracy, while 30 iterations adds marginal gains and serves as our default.

\section{Spurious Proof Example}
\label{app:spurious-proof}

Figure~\ref{fig:spurious_proof} shows a concrete example of a spurious proof observed under the No Rewriting setting.
The proof closes by \texttt{rfl} because the \texttt{abbrev solution} was defined to be literally the same set as the one appearing in the theorem statement — the prover never needed to reason about the underlying mathematics.

\begin{figure*}[t]
\centering
\begin{minted}{lean4}
abbrev fimo_2009_algebra_p3_solution : Set (ℕ+ → ℕ+) :=
  { f : ℕ+ → ℕ+ | (∀ x y,
      x + f y > f (y + f x - 1) ∧
      x + f (y + f x - 1) > f y ∧
      f y + f (y + f x - 1) > x) }

theorem fimo_2009_algebra_p3 :
  { f : ℕ+ → ℕ+ | (∀ x y,
      x + f y > f (y + f x - 1) ∧
      x + f (y + f x - 1) > f y ∧
      f y + f (y + f x - 1) > x) } =
    fimo_2009_algebra_p3_solution := by
  have h_main : { f : ℕ+ → ℕ+ | (∀ x y,
      x + f y > f (y + f x - 1) ∧
      x + f (y + f x - 1) > f y ∧
      f y + f (y + f x - 1) > x) } =
    fimo_2009_algebra_p3_solution := by rfl
  exact h_main
\end{minted}
\caption{A spurious proof of \texttt{fimo\_2009\_algebra\_p3} observed under the No Rewriting setting.
The \texttt{abbrev solution} is defined as the same set that appears in the theorem statement, so the prover closes the goal with \texttt{rfl} without performing any mathematical reasoning.}
\label{fig:spurious_proof}
\end{figure*}

\section{Cross-Model Pairing}
\label{app:cross-model}

A practical advantage of \method{}'s modular design is that the Discovery Module and the Proving Module can be replaced independently.
To demonstrate this flexibility, Table~\ref{tab:cross-model} reports results for three model combinations: our default pairing, a lightweight pairing using smaller open-source models, and a high-resource pairing using a stronger closed-source model.
Because the Aristotle API~\cite{achim2025aristotle} does not support concurrent requests and each problem can take hours, we sampled five problems per dataset for that condition; results are reported as solved/total.

\begin{table}[ht]
    \centering
    \caption{Pass@32 results for different Discovery–Proving model pairings.
    $^\dagger$Aristotle API results are sampled (5 problems per dataset).}
    \label{tab:cross-model}
\resizebox{\linewidth}{!}{%
    \begin{tabular}{lcccc}
      \toprule
      \raisebox{5pt}{\textbf{Configuration}} & \textbf{\shortstack{Putnam\\(660)}} & \textbf{\shortstack{Combi\\(100)}} & \textbf{\shortstack{mF2F\\(244)}} & \textbf{\shortstack{FIMO\\(149)}} \\
      \midrule
      \shortstack[l]{\footnotesize GPT-OSS 120B +\\ \footnotesize Goedel-V2 32B (Ours)} & \textbf{36} & \textbf{9} & \textbf{201} & \textbf{3} \\
      \shortstack[l]{\footnotesize Qwen3 8B +\\ \footnotesize DS-Prover-V1.5 7B}        & 5  & 2  & 107 & 2 \\
      \shortstack[l]{\footnotesize GPT-5 (Thinking) +\\ \footnotesize Aristotle API$^\dagger$} & 3/5 & 3/5 & 5/5 & 2/5 \\
      \bottomrule
    \end{tabular}}
\end{table}

Two conclusions follow.
First, the pipeline is functional even with small, resource-efficient models, supporting use in compute-constrained settings.
Second, replacing the Discovery Module with a significantly stronger model yields meaningful further gains, illustrating the headroom available as informal reasoning models continue to improve.

\end{document}